\documentclass[conference]{IEEEtran}
\IEEEoverridecommandlockouts
\usepackage{titlesec}					%custom \section
\usepackage{amssymb,amsmath}
\DeclareMathOperator{\sign}{sign}

\usepackage{algorithm}
\usepackage{algorithmic}
\usepackage{wrapfig}
\usepackage[sort&compress,numbers,sectionbib]{natbib}
\usepackage{cite} % Compacts [1]--[4] citations and adds spaces
\usepackage{booktabs}
\usepackage[small]{caption}
\usepackage{subcaption}
\usepackage{graphicx}

\usepackage{verbatimbox}
\usepackage{multirow}
\usepackage{balance}

\usepackage{booktabs}
\usepackage{tabularx}
\usepackage{subcaption}
\usepackage{caption}%[font=footnotesize]
\usepackage{wrapfig}
\usepackage{enumitem} % for alhabetic enumeration
\usepackage{xspace}
\usepackage{bm}
\usepackage{bbm} % bold math

\newcommand{\figref}[1]{Fig.~\ref{#1}}
\newcommand{\tabref}[1]{Table~\ref{#1}}
\usepackage[hyphens]{url}

\usepackage{hyperref}
\hypersetup{bookmarksopen,bookmarksnumbered,
pdfpagemode=UseOutlines,
colorlinks=true,
linkcolor=teal,
anchorcolor=teal,
citecolor=teal,
filecolor=teal,
menucolor=teal,
urlcolor=teal,
breaklinks=true
}

\usepackage[dvipsnames]{xcolor}
\definecolor{myred}{RGB}{215,48,39}
\definecolor{myblue}{RGB}{69,117,180}
\definecolor{myorange}{RGB}{252,141,89}
\definecolor{mylightblue}{RGB}{145,191,219}

\definecolor{MYlightblue}{RGB}{217,95,2}
\definecolor{MYdarkblue}{RGB}{117,112,179}
\definecolor{MYgreen}{RGB}{27,158,119}

\usepackage{color,soul}
\usepackage{svg}

\usepackage[normalem]{ulem}
\useunder{\uline}{\ul}{}

\usepackage{bm}


\usepackage{pgfgantt}

\def\BibTeX{{\rm B\kern-.05em{\sc i\kern-.025em b}\kern-.08em
    T\kern-.1667em\lower.7ex\hbox{E}\kern-.125emX}}
\begin{document}

\title{Implicit Communication in Human-Robot Collaborative Transport}

\author{
\IEEEauthorblockN{Elvin Yang}
\IEEEauthorblockA{
    \textit{Department of Robotics} \\
    \textit{University of Michigan} \\
    Ann Arbor, MI, USA \\
    eyy@umich.edu
}
\and
\IEEEauthorblockN{Christoforos Mavrogiannis}
\IEEEauthorblockA{
    \textit{Department of Robotics} \\
    \textit{University of Michigan} \\
    Ann Arbor, MI, USA \\
    cmavro@umich.edu
}
}

\maketitle

\begin{abstract}
We focus on human-robot collaborative transport, in which a robot and a user collaboratively move an object to a goal pose. In the absence of explicit communication, this problem is challenging because it demands tight \emph{implicit} coordination between two heterogeneous agents, who have very different sensing, actuation, and reasoning capabilities.
Our key insight is that the two agents can coordinate fluently by encoding subtle, communicative signals into actions that affect the state of the transported object.
To this end, we design an inference mechanism that probabilistically maps observations of joint actions executed by the two agents to a set of joint strategies of workspace traversal. Based on this mechanism, we define a cost representing the human's uncertainty over the unfolding traversal strategy and introduce it into a model predictive controller that balances between uncertainty minimization and efficiency maximization.
We deploy our framework on a mobile manipulator (Hello Robot Stretch) and evaluate it in a within-subjects lab study ($N=24$).
We show that our framework enables greater team performance and empowers the robot to be perceived as a significantly more fluent and competent partner compared to baselines lacking a communicative mechanism.
\end{abstract}

% HRI2025 requests that we use semicolons as delimiters instead of commas.

\section{Introduction}\label{sec:introduction}

Recently, there has been vivid interest in developing physically capable robot partners that could assist humans in \emph{context}-rich, dynamic and unstructured domains~\citep{selvaggio2021survey} like homes~\citep{homerobotovmmchallenge2023, van2011robocup} and manufacturing sites~\citep{matheson2019human}. An important task in this space involves the \emph{collaborative transport} of objects that might be too large or too heavy to be transported by a single agent. This task is especially challenging as it not only requires efficient and fluent coordination between the two heterogeneous partners but also the simultaneous satisfaction of geometric, kinematic, and physics constraints.

\begin{figure}[t]
  \includegraphics[width=\columnwidth]{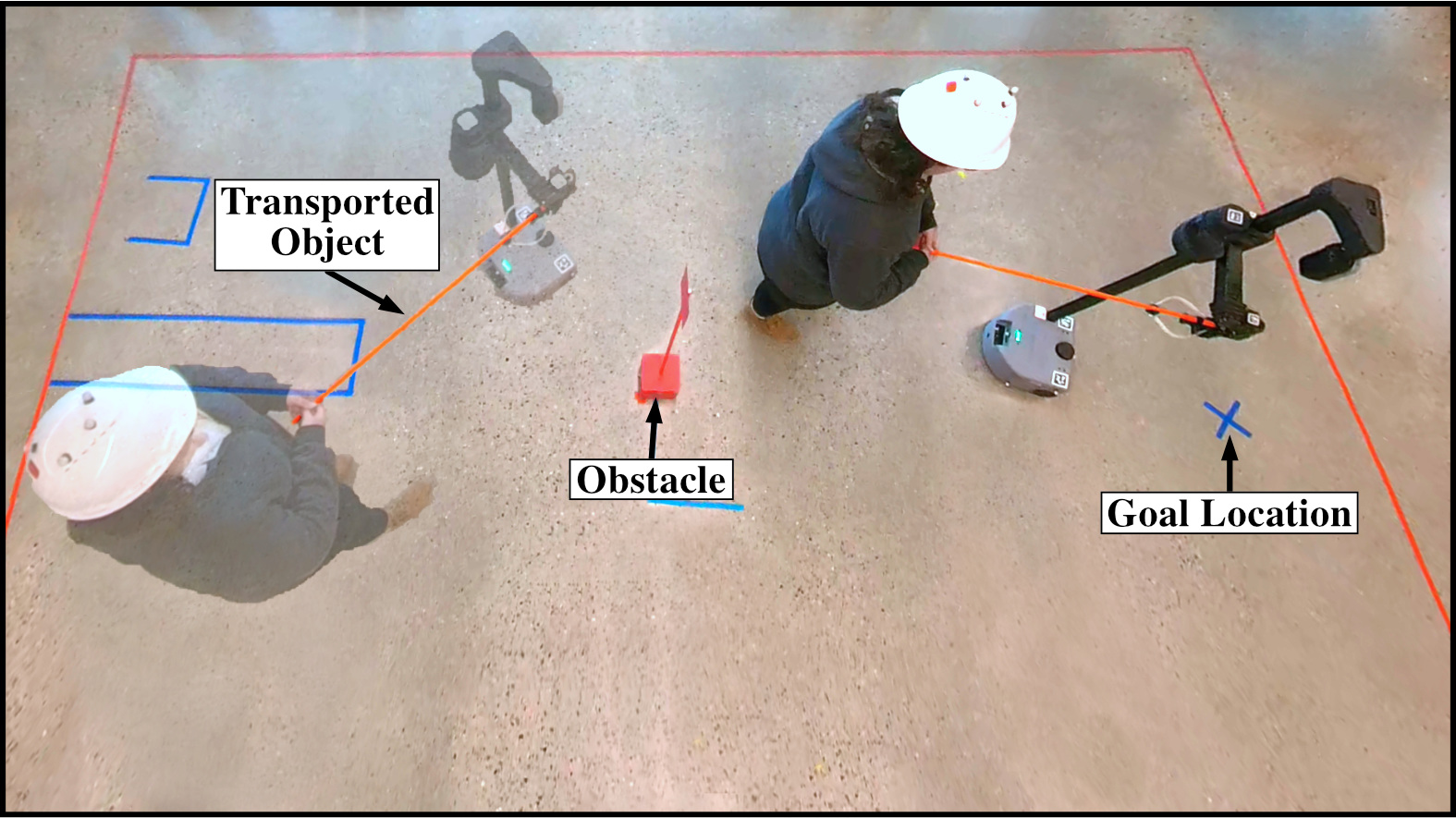}
  \caption{Footage from our study ($N=24$) involving the collaborative transport of an object (orange stick) by a user and a mobile manipulator in a workspace with an obstacle (red color). The robot runs our controller (IC-MPC), designed to balance functional and communicative actions in collaborative tasks.}
  \label{fig:teaser}
\end{figure}

Humans often tackle physically demanding collaborative tasks like transport by fluently coordinating their physical movements with their partners~\citep{sebanz2006joint} even without a concrete plan, with minimal explicit coordination. This capability relies on sophisticated mechanisms connecting perception and action. A prevalent theory from action understanding, commonly referred to as the ``teleological stance'', highlights that agents' actions can often be explained by an underlying goal~\citep{csibra2007obsessed,gergely1995intentional,baker2009understanding}. This idea has inspired researchers in human-robot interaction (HRI) to develop mechanisms that communicate a robot's intended goal to an observer through its actions~\citep{DraganAuR14,knepper2017implicit}. These mechanisms have produced intent-expressive robot behavior in manipulation~\citep{DraganAuR14}, autonomous driving~\citep{sadigh2016cars}, and social robot navigation~\citep{mavrogiannis2022socialmomentum}.
Likewise, we view communication---especially implicit communication~\citep{knepper2017implicit}, the ability to infer and convey information within physical actions---to be a critical skill of robots working in close physical collaboration with humans. Implicit communication can be low-latency, robust to environmental disturbances (e.g., noise, poor lighting conditions), and require less attention compared to explicit forms. While explicit communication remains highly relevant to team activities, implicit communication serves as an important complement that supports fluent teamwork.

To investigate the implications of implicit communication for physical human-robot teamwork, we instantiate a task of human-robot collaborative transport, where the goal of the human-robot team is to collaboratively move an object to a goal pose while avoiding collisions with static obstacles (see~\figref{fig:teaser}).
In this task, the user is simultaneously an observer of the robot and a \emph{dynamic actor}, persistently influencing and being influenced by the robot while it \emph{physically collaborates} with them.
While prior work in human-robot collaborative transport has emphasized fixed leadership roles for the two agents~\citep{bussy2012transportation,bussy2012proactive,mielke2024comanipulation,lima2023assistive,solanes2018human,nikolaidis2017mutual}, we consider a \emph{dynamic negotiation} over a joint strategy of \emph{workspace traversal}.
We contribute a control framework that leverages \emph{implicit communication}~\citep{knepper2017implicit} through actions influencing the state of the transported object to enable the robot to negotiate an efficient traversal with its human partner. We move beyond past work on implicit communication, where the user is either \emph{not an actor}~\citep{DraganAuR14} or \emph{not physically} collaborating with their robot partner~\citep{mavrogiannis2022socialmomentum,liang2019implicit}.
We demonstrate our framework on a mobile manipulator and evaluate it in a lab study ($N=24$) involving the collaborative transport of an object in a workspace with an obstacle obstruction. We show that our framework outperforms baselines lacking a communicative mechanism in terms of task completion and human impressions.

In summary, we contribute:
\begin{itemize}
  \item A formal mathematical representation of workspace traversal strategies during collaborative object transport.
  \item A human-inspired inference mechanism that probabilistically maps a joint human-robot action to a joint strategy of workspace traversal.
  \item A model predictive control framework that balances efficiency maximization and uncertainty minimization to produce fluent, efficient teamwork in physically collaborative tasks.
  \item Evidence from an extensive lab study ($N=24$) suggesting that our framework results in greater team performance and positively perceived robot behaviors. Videos from the study can be found at \url{https://youtu.be/0NTDrobSifg}.
  \item Code and data from our study that could help the community iterate on our work, publicly available at \url{https://github.com/fluentrobotics/icmpc\_collab\_transport}.
\end{itemize}

\section{Related Work}\label{sec:related-work}

We discuss relevant work on human-robot collaboration (HRC), our target domain of human-robot collaborative transport, and our technical foundation of implicit communication.

\subsection{Human-Robot Collaboration}

HRC, defined broadly as the integration of humans and robots working together towards a set of (possibly joint) goals, has attracted considerable attention in recent years~\citep{mainprice2013collaborative,Wilcox-RSS-12,gopinath2017sharedautonomy,carlson2013wheelchair-shared}. An important challenge in HRC involves determining a fluent meshing between human and robot actions that leverages the unique competences of both in an effective way~\citep{hoffman2019evaluating}. This has motivated research on the design of planning and control frameworks that smoothly blend human and robot control inputs~\citep{dragan2013blending,nikolaidis2017adaptation,unhelkar2020bidirectional,aronson2024intentional}. For many applications, the human and the robot goals are well-defined; e.g., in shared control for wheelchair navigation, the role of the robot is to lead the user to a desired waypoint~\citep{carlson2013wheelchair-shared}. However, in other applications, the richness of the context results in a multiplicity of ways in which a task could be achieved. In those cases, a robot has to reason about how its own strategy for satisfying its objectives meshes with the strategy of its human partner~\citep{zhao2022coordination,pandya2024multi,peters2020inference,wang2022co} or how its actions contribute towards a joint strategy~\citep{xie2021learning}. An underlying challenge involves determining an appropriate space of strategies. \citet{nikolaidis2013human} describe an interactive planning framework that iteratively switches role assignment to humans and robots with the goal of converging on a shared strategy. \citet{zhao2022coordination} and \citet{xie2021learning} learn spaces of strategies from observations of low-level action sequences whereas \citet{wang2022co} learn a latent space of strategies that enables a robot to adapt to its collaborator.

Our approach is motivated by the physical HRC task of human-robot collaborative transport. The execution of this task admits multiple possible solutions in realistic environments with obstacle obstructions.
Recent work in HRC emphasizes the extraction of these solutions from team strategies observed in human teamwork. In contrast, we exploit the mathematical structure of the domain, identifying strategies of collaborative transport as classes of homotopy-equivalent trajectories of workspace traversal~\citep{knepper2012equivalence,vernaza2013winding,kretzschmar2016irl}. This allows for methodical enumeration of strategies in an interpretable form and in line with insights from recent studies on human-human object transport~\citep{freeman2023classification}. Additionally, while a body of work considers validation in virtual benchmarks, we emphasize \emph{physical}, \emph{explicit}, and \emph{concurrent} human-robot collaboration~\citep{selvaggio2021survey} treating the robot as a fully embodied partner with mobile manipulation capabilities.

\subsection{Human-Robot Collaborative Transport}

Physical HRC can support applications like personal robotics, manufacturing, and construction~\citep{losey2018review-shared,ogenyi2019physical,selvaggio2021survey}. Much of the recent work has focused on collaborative manipulation tasks involving a user and a manipulator, such as the rearrangement of large or heavy objects~\citep{gienger2018contactchanges,kim2018anticipatory,stouraitis2020bilevel,al2023resolving,kucukyilmaz2019online,agravante2014collaborative,Zheng2022SafeHC,Shao2024comanipulation} and the collaborative use of tools for sawing or bolt screwing~\citep{peternel2017comanipulation}.

Many real-world applications motivate the integration of manipulation and robot mobility. An important task of interest is the \emph{collaborative transport} of an object by a user and a mobile manipulator. This task is especially challenging because in addition to human-robot coordination, it requires considerations like collision avoidance, object stability, and human ergonomics. A large body of work assumes a fixed role assignment: the human intent is estimated and the robot follows it~\citep{bussy2012transportation,bussy2012proactive,mielke2024comanipulation,lima2023assistive,solanes2018human}. This can be too restrictive, especially when the robot has the capability to contribute meaningfully to the task as an independent agent. To account for such situations, \citet{nikolaidis2017mutual} develop a probabilistic controller that estimates the user's preferences and generates corrective actions to guide them to an efficient strategy. However, this implies that the robot is more equipped to guide the interaction, which might not always be the case. Instead, \citet{ng2023takes} allow leadership to be an emergent property of following a policy extracted from human demonstrations. \citet{mortl2012role} engineer a dynamic leadership negotiation through actions executed on the transported object by the two agents.
While mobility is integrated across these works, it is often oversimplified through the assumption of predefined paths~\citep{mortl2012role} or collision-free workspaces~\citep{bussy2012transportation,bussy2012proactive}.

In this work, we leverage the full mobility of a mobile base and explicitly account for obstacles while negotiating a strategy of workspace traversal. We develop a control framework that enables the robot to flexibly negotiate with the user using implicit communication, realized via subtle signals encoded in velocities transmitted to the transported object. In contrast to approaches directly learning traversal policies from human demonstrations~\citep{ng2023takes}, our approach has the potential to handle diverse maps by reasoning about the mapping between team actions and traversal strategies. While prior work assumes heavily instrumented setups including sensorized objects~\citep{mielke2024comanipulation,bussy2012proactive,bussy2012transportation,gienger2018contactchanges}, we use a commodity mobile manipulator (Hello Robot Stretch) and limited additional instrumentation (a motion capture system to eliminate the influence of localization errors). Finally, we move beyond prior work in terms of validation depth by conducting an extensive lab study ($N=24$) with in-depth insights via objective and subjective measures.

\subsection{Implicit Communication}

Any deviation from a robot's expected behavior has the potential to convey important, task-related information to an observer~\citep{knepper2017implicit}. This idea, rooted in conversational implicature~\citep{grice1975logic} has deep relevance to HRI~\citep{urakami2023nonverbal,hoffman2024inferring,venture2019expressive}. The community has long been exploring the expressive power of modalities like motion~\citep{takayama2011expressing}, gaze~\citep{admoni2014deliberate}, or gestures~\citep{carter2014catch}. Researchers have been building algorithmic frameworks that harness implicit communication to engineer fluent human-robot teamwork. \citet{knepper2017implicit} formalize implicit communication as the minimization of the information entropy over a probability distribution of possible goals. Considering a social navigation domain, \citet{mavrogiannis2022socialmomentum} implement implicit communication into a navigation controller that conveys the robot's intended passing side to co-navigating users. Considering a coverage task, \citet{walker2021influencing} implement implicit communication into the generation of robot motion that influences an observer to summarize robot motion in a desirable way. In a manipulation domain, \citet{DraganAuR14} instantiate implicit communication as \emph{Legibility}, a property of robot motion that enables an observer to infer the target object of a manipulator's reaching motion. Considering the collaborative game Hanabi, \citet{liang2019implicit} demonstrate that implicit communication can be a powerful tool for teams of humans and artificial agents working on a joint task.

Prior algorithmic instantiations of implicit communication assume simplified teamwork settings. For instance, the user is often passively observing the robot~\citep{DraganAuR14,walker2021influencing}, navigating next to the robot~\citep{mavrogiannis2022socialmomentum}, or working with a non-embodied agent~\citep{liang2019implicit}. In contrast, this work integrates implicit communication into physical human-robot collaboration through the task of collaborative transport. This task uniquely integrates the user and the robot as embodied, dynamic observers and actors, whose actions influence each other and impact the quality of the teamwork. We define a space of goals, corresponding to strategies of workspace traversal, which we identify using a notion of topological invariance~\citep{mavrogiannis2023winding}. We then develop a probabilistic inference that maps joint human-robot actions to strategies of workspace traversal. By tracking the entropy of the distribution over strategies, the robot may monitor the state of uncertainty over the unfolding traversal. By introducing entropy as a cost into a model predictive controller (IC-MPC), the robot is capable of balancing functional and communicative actions resulting in efficient, fluent, and positively perceived human-robot teamwork.

\section{Problem Statement}\label{sec:problem-statement}

\begin{figure}[t]
  \includegraphics[width=\columnwidth]{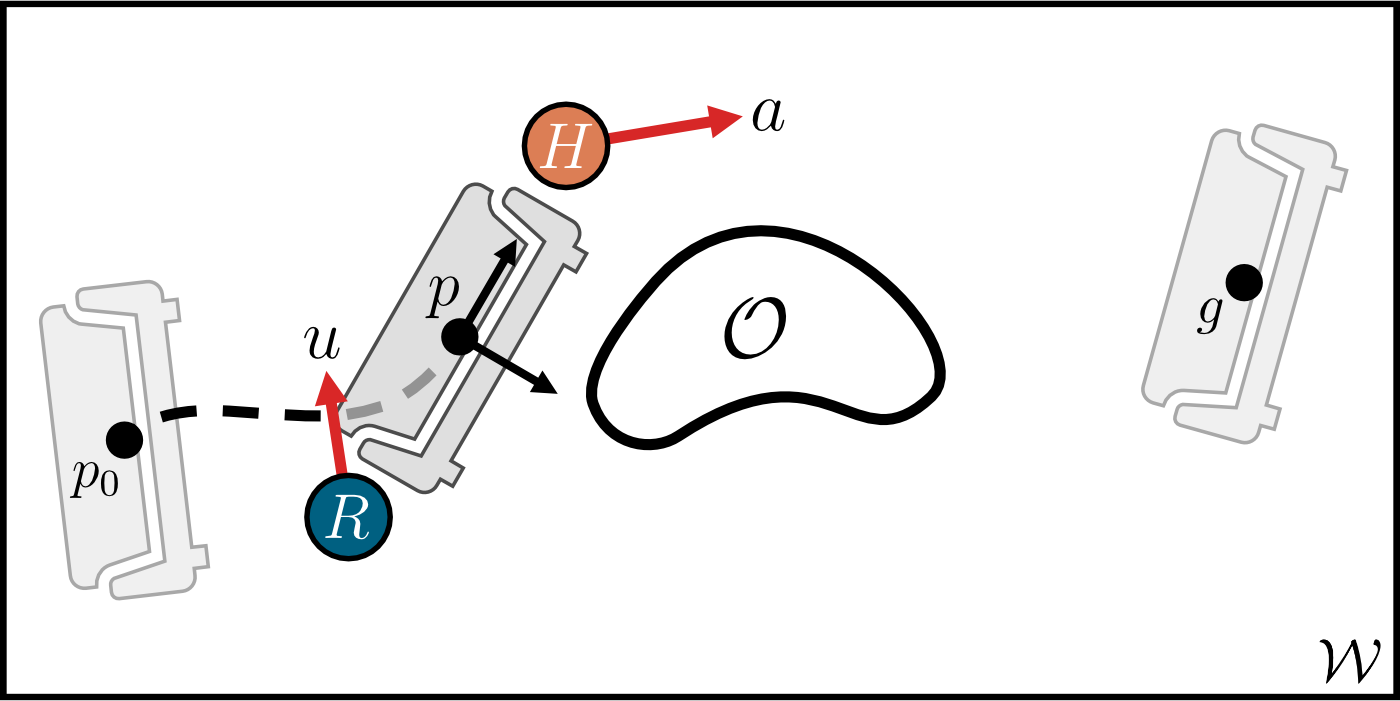}
  \caption{A human (H) and a robot (R) collaboratively move an object from an initial pose $p_0$ to a final pose $g$ in a workspace $\mathcal{W}$. An obstacle $\mathcal{O}$ stands in their way. To avoid collisions with $\mathcal{O}$ and reach $g$, they have to coordinate on a strategy of workspace traversal. In this work, we engineer implicit coordination through the velocities $a$ and $u$ that the human and the robot exert on the object.
  }
  \label{fig:setup}
\end{figure}

We consider a human $H$ and a robot $R$ collaboratively transporting an object. The robot and the human grasp the object at a fixed height; this allows us to instantiate the problem on a planar workspace $\mathcal{W}\subseteq SE(2)$.
Assuming a quasistatic setting, the object's state $p \in \mathcal{W}$ evolves according to $p_{k+1} = f(p_k, a_k, u_k)$, where $a\in\mathcal{A}$, $u\in\mathcal{U}$ represent human and robot velocities, respectively, and $k$ is a time index.
The workspace includes a set of obstacle-occupied regions $\mathcal{O}\subset\mathcal{W}$. The goal of the human-robot team is to transport the object from an initial pose $p_0$ to a desired pose $g$ in $\mathcal{W}$ (see~\figref{fig:setup}) while avoiding collisions with $\mathcal{O}$. We assume that the two agents do not communicate explicitly (e.g., via language), but they observe the actions of one another. Our goal is to design a control policy to enable the robot to efficiently and fluently collaborate with its human partner.

\section{Balancing Functional and Communicative Actions in Human-Robot Collaborative Transport}

We describe a control framework that leverages implicit communication to support efficient and fluent collaboration in human-robot collaborative transport. By reasoning about its partner's uncertainty over the \emph{way} the task is being executed, the robot balances between communicative, uncertainty-reducing actions, and functional, task-driven actions. This balance is not prescribed, but rather dynamically adaptive to the robot's belief about the uncertainty of its partner.

\subsection{Formalizing Joint Strategies of Workspace Traversal}

Collaborative tasks involving multiple agents working together require consensus on a \emph{joint strategy} $\psi$, i.e., a qualitatively distinct way of completing the task, out of the set of all possible joint strategies, $\Psi$. Often, this consensus is not established \textit{a priori}; rather, it is dynamically negotiated during execution. The abstraction of a joint strategy effectively captures critical domain knowledge at a representation level. While prior work on collaborative transport has emphasized \emph{role} assignment across the team (i.e., whether the robot or the human are leading or following each other)~\citep{mortl2012role, nikolaidis2013human, jarrasse2014slaves}, realistic, obstacle-cluttered environments present additional important challenges, such as the decision over \emph{how to pass through} an obstacle-cluttered workspace.

In this work, we formalize the space of workspace traversal strategies using tools from homotopy theory~\citep{knepper2012equivalence}. The human-robot team is tasked with transporting an object from its initial pose $p_0$ to a final pose $g$, resulting in an object trajectory $\boldsymbol{p}:[0,1]\to\mathcal{W}$, where $\boldsymbol{p}(0)=p_0$ and $\boldsymbol{p}(1)=g$, belonging to an appropriate space of trajectories $\mathcal{P}$.
Obstacles, defined as the connected components of $\mathcal{O}$, naturally partition $\mathcal{P}$ into equivalence classes $\Psi$, where each $\psi\in\Psi$ represents a distinct workspace traversal strategy under which the transported object can travel from $p_0$ to $g$, i.e.,
\begin{equation}
    \begin{split}
    & \mathcal{P} = \underset{\psi \in \Psi}{\bigcup} \psi \\
    & \forall \psi^i, \psi^j \in \Psi : (\psi^i \cap \psi^j \neq \emptyset) \implies (\psi^i = \psi^j) \\
    & \forall \boldsymbol{p}^i, \boldsymbol{p}^j \in \psi : \boldsymbol{p}^i \sim \boldsymbol{p}^j
    \end{split}
    \mbox{.}
\end{equation}

\begin{figure}[t]
\begin{subfigure}{\linewidth}
    \centering
    \includegraphics[width=\linewidth]{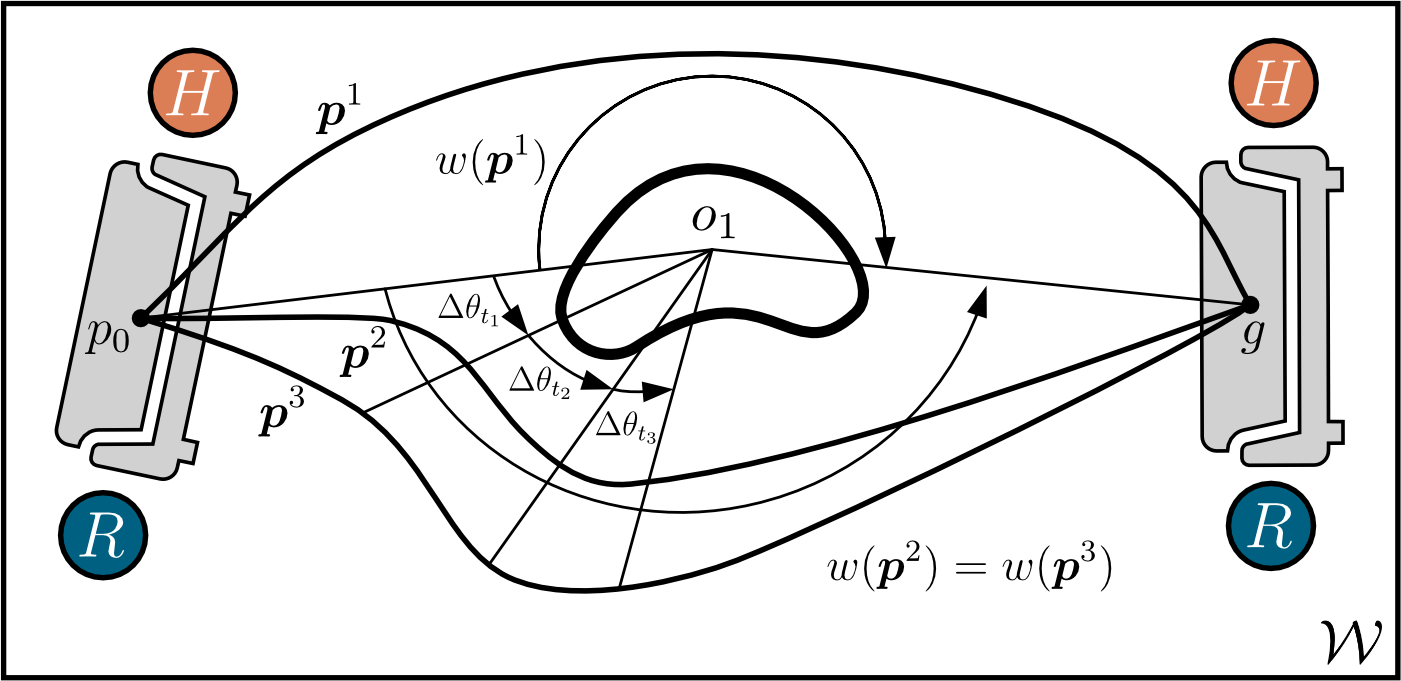}
    \caption{
    Identification of workspace traversal strategies based on path homotopy~\citep{kretzschmar2016irl}. By integrating the angle of the vector between the obstacle and the object as it is being transported along a path $\boldsymbol{p}$, we extract a winding number $w(\boldsymbol{p})$ identifying the strategy of workspace traversal. Here, $w(\boldsymbol{p}^2)=w(\boldsymbol{p}^3)$ since both $\boldsymbol{p}^2$, $\boldsymbol{p}^3$ passed on the right of $o_1$.}
    \label{fig:winding}
\end{subfigure}
\begin{subfigure}{\linewidth}
    \includegraphics[width=\linewidth]{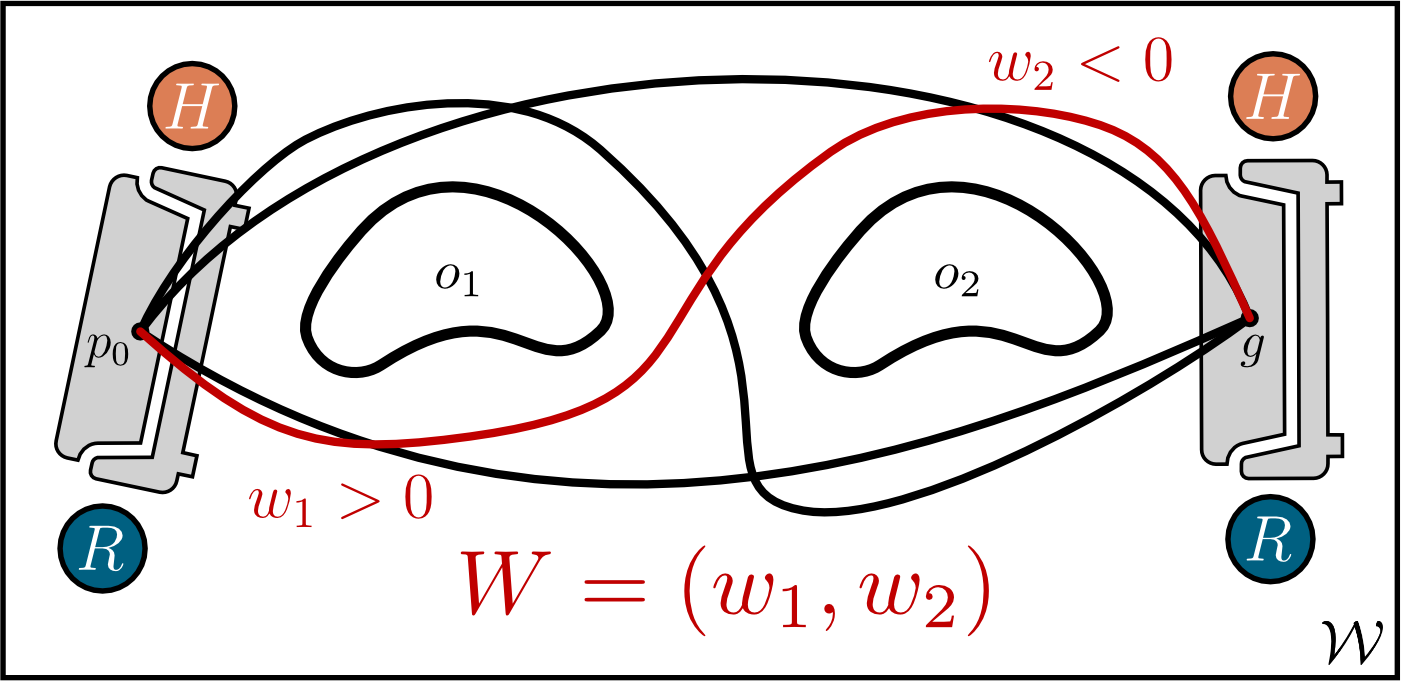}
    \caption{Representing workspace traversal strategies as tuples of winding number signs, $W$. In this scene with two obstacles, there are four possible strategies represented as continuous curves. The red curve highlights a strategy corresponding to passing on the right of $o_1$ ($w_1>0$), and the left of $o_2$ ($w_2<0$). This representation is applicable to any number of obstacles.}
    \label{fig:passing-strategy}
\end{subfigure}
\caption{Illustration of our topological abstraction for representing strategies of workspace traversal.\label{fig:workspace-traversal}}
\end{figure}

These classes can be identified using a notion of topological invariance. The works of~\citet{vernaza2013winding,kretzschmar2016irl,mavrogiannis2023winding} use winding numbers to describe topological relationships between the robot and obstacles or humans navigating around it. Here, we adapt this idea to collaborative transport by enumerating the set of homotopy classes between the object trajectory and obstacles in the workspace. Specifically, for any object trajectory $\boldsymbol{p}$ embedded in a space with $m$ obstacles $o_1,\dots, o_m$, we can define winding numbers
\begin{equation}
    w_i = \frac{1}{2\pi}\sum_{t} \Delta\theta_t^i, \quad i=1,\dots, m\mbox{,}
\end{equation}
where $\Delta\theta^i_t = \angle \left(p_t - o_i, p_{t-1}-o_i\right)$ denotes an angular displacement corresponding to the transfer of the object from $p_{t-1}$ to $p_t$ (see~\figref{fig:winding}). The sign of $w_i$ represents the passing side between the object and the $i$-th obstacle, and its absolute value represents the number of times the object encircled the $i$-th obstacle.
For a trajectory $\boldsymbol{p}$, the tuple of winding number signs
\begin{equation}
    W=(\sign{w_1},\dots, \sign{w_m})
\end{equation}
represents an equivalence class describing how the human-robot team transported the object past all obstacles in the environment. In this work, we model the space of joint strategies $\Psi$ as set of distinct $W$, i.e., $|\Psi|=2^m$.

\subsection{Inferring Strategies of Workspace Traversal}

We describe an inference mechanism that maps observations of team actions to a belief over a workspace traversal strategy. This mechanism is agnostic to the specific definition of the strategy. At time $t$, we assume that the robot observes the joint action $\alpha = (a, u)$, the object state $p$, and the task context $c = (g, \mathcal{O})$. Given $\alpha$, $p$, and $c$, our goal is to infer the unfolding workspace traversal strategy, $\psi$, i.e.,
\begin{equation}
    \mathbb{P}(\psi \mid \alpha, p, c)\mbox{.}\label{eq:inference}
\end{equation}
Using Bayes' rule, we can expand~\eqref{eq:inference} as
\begin{equation}
    \mathbb{P}(\psi \mid \alpha, p, c) = \frac{1}{\eta} \mathbb{P}(\alpha \mid \psi, p, c)\,\mathbb{P}(\psi \mid p, c)\mbox{,}\label{eq:inference-bayes}
\end{equation}
where the left-hand side expression is the \emph{posterior distribution} of the joint strategy $\psi$, and on the right-hand side, $\eta$ is a normalizer across $\alpha$, $\mathbb{P}(\alpha \mid \psi, p, c)$ is the \emph{joint action likelihood distribution} and $\mathbb{P}(\psi \mid p, c)$ is a \emph{prior distribution} of the joint strategy before observing the joint action. We can rewrite the joint action likelihood distribution as
\begin{equation}
    \mathbb{P}(\alpha \mid \psi, p, c) = \mathbb{P}(a \mid \psi, p, c)\,\mathbb{P}(u \mid \psi, p, c)\mbox{,}
\end{equation}
since the two agents choose their actions independently.

The distribution of \eqref{eq:inference} allows the robot to represent the belief of its partner over the unfolding traversal strategy. A natural measure of uncertainty over the observer's belief regarding that strategy can be acquired by computing the information entropy of $\psi$, conditioned on known $\alpha, p, c$:
\begin{equation}
    H\left(\psi \mid \alpha,p,c\right) = - \sum_{\psi \in \Psi} \mathbb{P}(\psi\mid\alpha,p,c) \log \mathbb{P}(\psi\mid\alpha,p,c)\label{eq:entropy}    \mbox{.}
\end{equation}
Intuitively, the higher $H$ is, the higher the uncertainty of the user over the unfolding $\psi$ is assumed to be.

\subsection{Integrating Human Inferences into Robot Control}\label{sec:control}

We integrate the inference mechanism of \eqref{eq:inference} into a model predictive control (MPC) algorithm by using its entropy \eqref{eq:entropy} as a cost. Given the context $c = (g, \mathcal{O})$ and the object state $p$ at time $t$, the goal of the MPC is to find the sequence of future robot actions $\boldsymbol{u}^*$ that minimizes a cost function $J$ over a horizon $T$. At every control cycle, the MPC solves the following planning problem:

\begin{equation}
\begin{split}
\left(u_{t:t+T}\right)^{*} = \underset{u_{t:t+T}} {\arg\,\min}\; & J(p_{t:t+T}, u_{t:t+T})\\
    s.t.\: & p_{k+1} = f(p_k, a_k, u_k) \\
           & a_k \in \mathcal{A} \\
           & u_k\in\mathcal{U}
   \label{eq:mpc}
\end{split}\mbox{,}
\end{equation}
We split $J$ into a running cost $J_k$ and a terminal cost $J_T$
\begin{equation}
\begin{split}
     J(p_{t:t+T}, u_{t:t+T}) = & \sum_{k=0}^{T} \gamma^k J_k(p_{t+k}, u_{t+k})\\
        &+ J_T(p_{t+T}, u_{t+T})
\end{split}
        \mbox{,}
\end{equation}
where $\gamma$ is a discount factor, and the terminal cost penalizes distance from the object's goal pose $g$:
\begin{equation}
    J_T(p_{t+k}, u_{t+k}) = || p_{t+k} - g ||^2
    \mbox{.}
\end{equation}
The running cost $J_k$ is a weighted sum of two terms, i.e.,
\begin{equation}
    \begin{split}
        J_k(p_{t+k}, u_{t+k}) =\,& w_{obs} J_{obs}(p_{t+k}, u_{t+k})\\ &+ w_{ent} J_{ent}(p_{t+k}, u_{t+k})\mbox{,}
    \end{split}\label{eq:mpc-running-cost}
\end{equation}
where
\begin{equation}
    \begin{split}
    J_{obs}(p_{t+k}, u_{t+k}) &=\\ \max & \left(0, -\log\left(\underset{o \in \mathcal{O}}{\min} \frac{||p_{t+k} - o||}{\delta}\right)\right)
    \end{split}
    \mbox{,}
\end{equation}
is a collision avoidance cost penalizing proximity to obstacles, $\delta$ is a clearance threshold, $J_{ent}$ is a cost proportional to the entropy defined in \eqref{eq:entropy}, and $w_{obs}$, $w_{ent}$ are weights.

We refer to this control framework as \emph{Implicit Communication MPC}, or \textbf{IC-MPC}. At every control cycle, IC-MPC plans a future robot trajectory that balances between functional objectives (collision avoidance, progress to goal) and communicative objectives (minimizing the partner's uncertainty over the upcoming joint strategy). The robot executes the first action $u_t$ from the planned trajectory and then replans. This process is repeated in fixed control cycles until the task is completed.

\section{User Study}\label{sec:evaluation}

We conducted an IRB-approved, within-subjects user study (U-M HUM00254044) in which a user collaborates with a robot to jointly transport an object to a designated pose. Each user experienced the same set of conditions, each corresponding to a collaborative algorithm running on the robot: ours, and two baselines. Through mailing lists, we recruited 24 participants (4 female, 18 male, 2 other), aged 18-29 from a university population. On average, participants rated of their familiarity with robotics technology as 4.1 (SD = 0.74) on a scale from 1 (not at all familiar) to 5 (very familiar). The study lasted 45 minutes, and each participant received \$20 of compensation.

\subsection{Experiment Design}

\begin{figure}[t]
    \centering
    \includegraphics[width=\columnwidth]{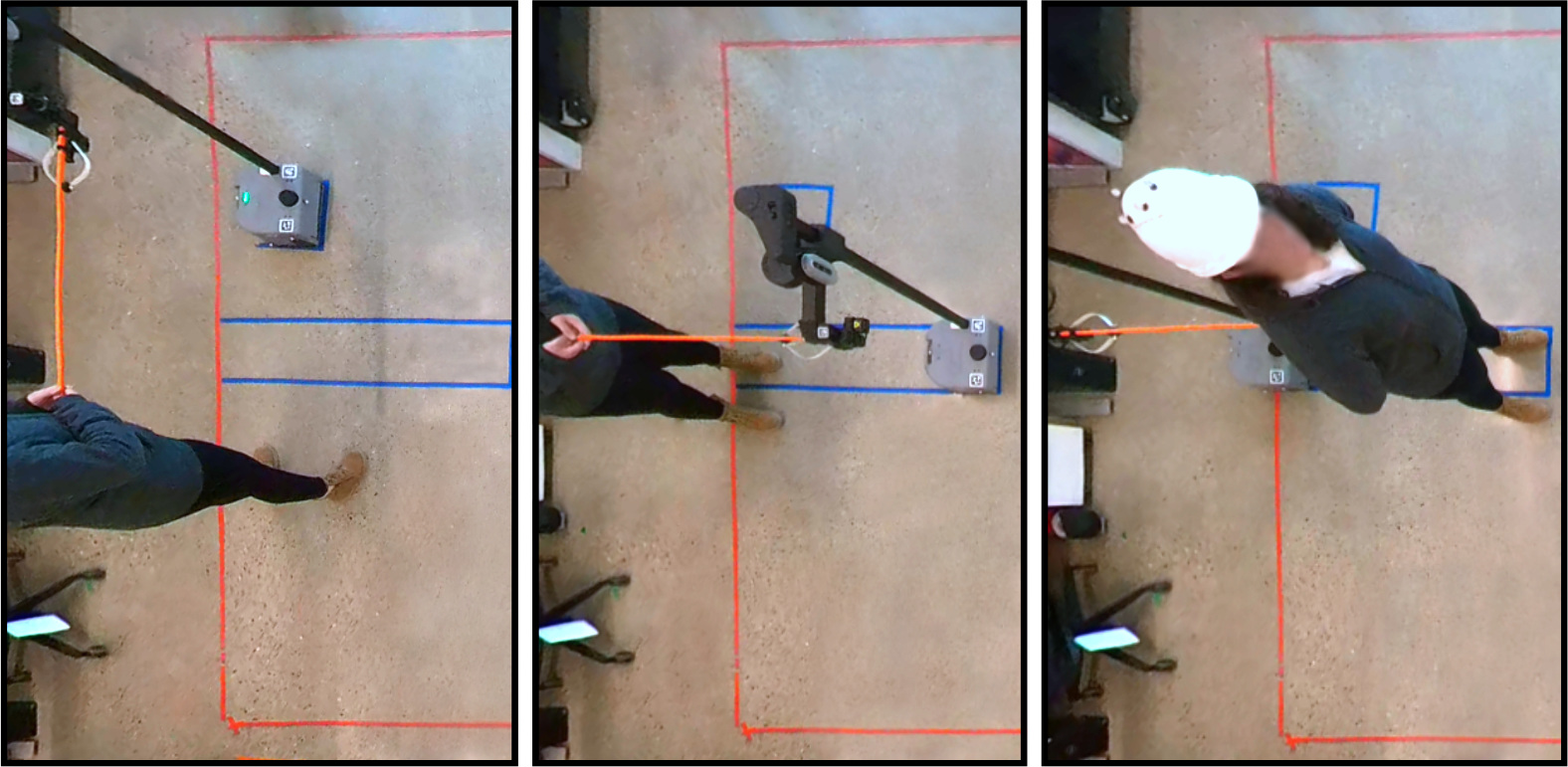}
    \caption{Users experienced three different starting configurations with each robot algorithm. From left to right: the user and the robot stand side-by-side; the user stands directly behind the robot; the user stands directly in front of the robot. For the third configuration, the user faces toward the goal and holds the object behind their back.}
    \label{fig:exp-setup}
\end{figure}

\textbf{Task Description.} The user and the robot hold opposite ends of an object (a wooden stick) and transport it together from an initial pose to a goal pose.
Users collaborate with each algorithm three times to ensure they experience a diverse range of interactions.
At the start of each of the three trials, the user and the robot stand in predetermined configurations (\figref{fig:exp-setup}), following the same fixed order for all algorithms.

\textbf{Algorithms.}
We compare the performance of our framework (\textbf{IC-MPC}) against two baselines:

\begin{itemize}
    \item \emph{Vanilla-MPC}: A purely functional ablation of IC-MPC with no uncertainty-minimizing objective ($w_{ent} = 0$).
    \item \emph{VRNN}~\citep{ng2023takes}: A learning-based path planner based on a Variational Recurrent Neural Network that predicts the most likely future path of the object based on human demonstrations. The robot takes actions to track path predictions as closely as possible.
\end{itemize}

\textbf{Metrics.} We evaluate performance in terms of:

\textit{Objective Metrics}:
\begin{itemize}
    \item \emph{Success rate}, defined as the proportion of successful trials. A trial is successful if the following conditions hold:
    a) any part of the object reaches the goal; b) the object stays within the workspace boundary; c) the object stays parallel to the ground---i.e., not lifted above the obstacle or dropped; d) there are no obstacle collisions.
    \item \emph{Completion time}, measured as the time taken for the object to reach the goal.
    \item \emph{Acceleration}, measured as the average acceleration of the human trajectory similar to prior work~\citep{mavrogiannis2022socialmomentum}.
\end{itemize}

\textit{Subjective Metrics}:
\begin{itemize}
    \item \emph{Warmth}, \emph{Competence}, and \emph{Discomfort}, measured using the RoSAS scale~\citep{carpinella2017robotic}.
    We use the original form of the scale: a single list of 18 items in randomized order and a nine-point response scale from 1 = ``Definitely not associated'' to 9 = ``Definitely associated''.
    \item \emph{Fluency}, measured using the \emph{Fluency in HRI} scale~\citep{hoffman2019evaluating}. To reduce fatigue, we use seven of eight items that were validated with objective measures in the original paper~\citep{hoffman2019evaluating}, with one item removed as recommended.
    The items are presented in randomized order with a seven-point response scale from 1 = ``Strongly disagree'' to 7 = ``Strongly agree''.
\end{itemize}

\textbf{Hypotheses.} Our insight is that implicit communication will have observable implications for the collaboration quality of the human-robot team, at an objective and subjective level. We formalize this insight into the following two hypotheses:

\begin{enumerate}[label=\textbf{H\arabic*}]
    \item IC-MPC is more effective at completing this task in collaboration with a user compared to Vanilla-MPC and VRNN as measured by the objective metrics.
    \item IC-MPC is viewed more favorably as a collaborator compared Vanilla-MPC and VRNN as measured by users' responses to questionnaires containing the subjective metrics.
\end{enumerate}

\subsection{System Development}

\textbf{Experimental Setup}. We deploy all algorithms on Stretch RE2 from Hello Robot~\citep{kemp2022design}, a mobile manipulator with a differential-drive mobile base and a prismatic arm.
To avoid excessive torque on the robot's end-effector, the transported object is a lightweight stick of $0.914~m$ length and $0.05~kg$ mass.
The robot's wrist motors are turned off to allow free rotation of the end effector. The lift of the robot arm is static throughout the task to constrain the object's movement to $SE(2)$.
Across our experiments, the team operates in a workspace with area $2.8 \times 5.6~m^2$.
To study the coordination of the human-robot team over a discrete decision, a single static obstacle of area $0.15 \times 0.15~m^2$ is placed in the center of the workspace (see~\figref{fig:teaser}). The workspace is fitted with an overhead Optitrack Flex 13 motion-capture (mocap) system that continuously streams poses and velocities of the robot and the user (via a construction-style helmet) at 120 Hz.

\textbf{Human modeling}. Since our study involved an environment with a single obstacle, we set $\Psi = \{\textsc{left}, \textsc{right}\}$, corresponding respectively to $w<0$ and $w>0$.
We instantiated the strategy inference using analytical models of the prior distribution and joint action likelihood distribution~\eqref{eq:inference-bayes}. However, it is possible to approximate this inference using data-driven techniques, e.g., by learning them from datasets of demonstrations~\citep{ng2023takes, freeman2023classification}.

We model the prior distribution over the joint strategy as a function of the winding number $w$. Without loss of generality, when $p_0, o, g$ are collinear, we consider the obstacle to have been passed when $\left|w\right| \ge \frac{1}{4}$:
\begin{equation}
\begin{split}
    \mathbb{P}(\textsc{left} \mid p, c) &= \max\left(0, \min\left(0.5 - 2w, 1\right)\right) \\
    \mathbb{P}(\textsc{right} \mid p, c) &= \max\left(0, \min\left(0.5 + 2w, 1\right)\right)
\end{split}
\end{equation}

\begin{figure}
    \centering
    \includegraphics[width=\columnwidth]{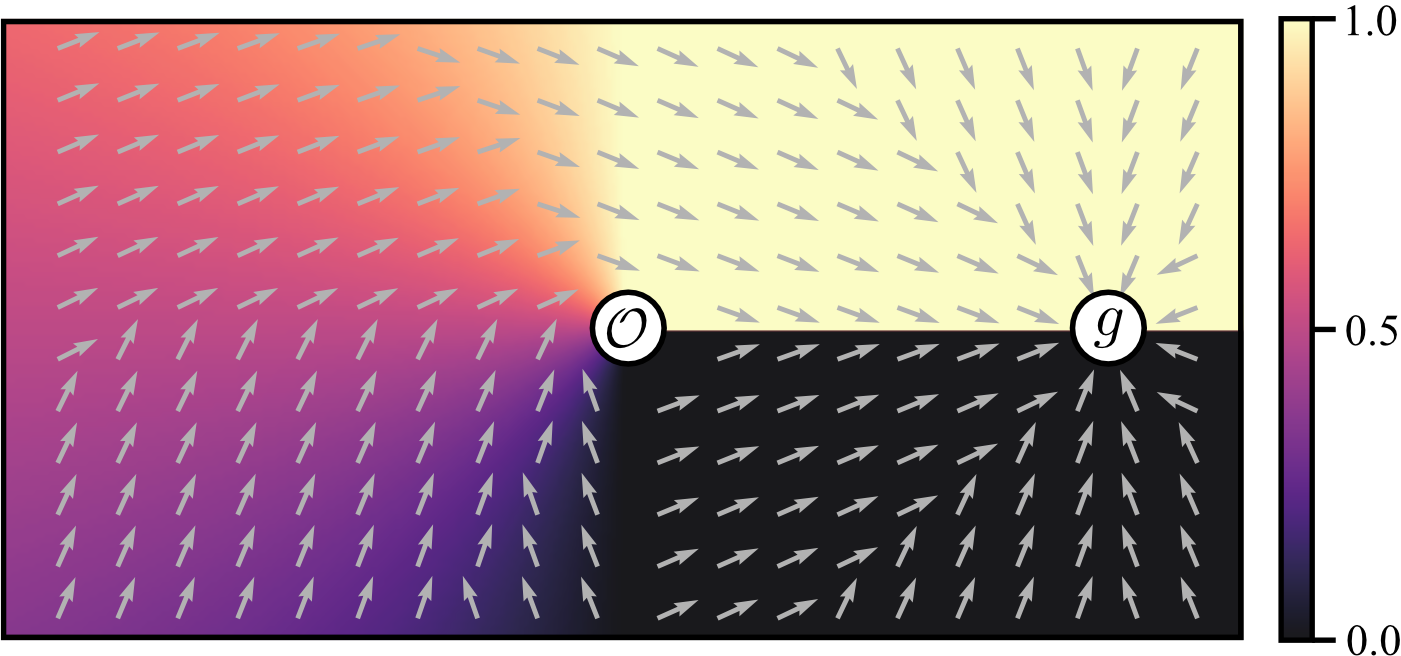}
    \caption{Workspace traversal strategy inference. The prior distribution for $\mathbb{P}(\textsc{left} \mid p, c)$ is shown as a colormap in the background, and the mode of the action likelihood distribution for $\mathbb{P}(a \mid \textsc{left}, p, c)$ is shown using gray arrows in the foreground.}
    \label{fig:prior-distribution}
\end{figure}

Before the obstacle is passed, the action likelihood distribution models the most likely action as the velocity that maximizes change in the winding number $w$. We approximate this as
\begin{equation}
\begin{split}
    \mathbb{P}(a \mid \textsc{left}, p, c) &\propto \exp\left(a \cdot R\left(\frac{\pi}{3}\right) \overrightarrow{po}\right) \\
    \mathbb{P}(a \mid \textsc{right}, p, c) &\propto \exp\left(a \cdot R\left(-\frac{\pi}{3}\right) \overrightarrow{po}\right)
\end{split}
\end{equation}
where $R(\cdot)$ is a 2D rotation matrix.
After the obstacle is passed, the most likely action is instead in the direction of the goal.
We illustrate the prior and action likelihood distributions for $\psi=\textsc{left}$ in \figref{fig:prior-distribution}. The same model of the action likelihood distribution is used for human and robot actions.

Prior to the user study, we evaluated our analytical models using the simulation dataset provided by~\citet{ng2023takes}.
We considered the subset of dataset environments with a single obstacle and annotated each trajectory with a ground truth joint strategy label.
The prior and action likelihood distributions achieved 98.9\% and 72.4\% accuracy in this subset of the dataset, respectively.

\textbf{MPC}.
All algorithms are implemented using an open-source package~\citep{arm-mppi} for Model Predictive Path Integral (MPPI) control~\citep{williams2017mppi}.
The same discount factor $\gamma = 0.95$, planning horizon $15 \times 0.25\,s$, and number of rollout samples $(100)$ is used across all algorithms. IC-MPC and Vanilla-MPC use obstacle clearance threshold $\delta = 0.5$ and weight $w_{obs} = 1.0$. IC-MPC uses entropy weight $w_{ent} = 1.0$.
These parameters were empirically selected to ensure good performance in our environment during pilot hardware trials, and different workspace configurations may require different parameter combinations.
Observations of human velocities are downsampled to 10 Hz from the motion capture, and a constant velocity model is used for the human motion prediction rollouts~\citep{scholler2020constant}.
All algorithms run on an Intel i7-13700 CPU at 15 Hz.

\textbf{VRNN}.
We adapt MPC to track VRNN's prediction of the object's future path as closely as possible as in~\citet{ng2023takes}. However, we found that path predictions occasionally have undefined orientations, e.g., the output of the model contains states for which $\left|\cos \theta\right| > 1$.
Consequently, we use only the translational component of the path predictions.
Since VRNN is designed to avoid obstacles and reach the goal, we use a running cost based on Euclidean distance to the predicted path and no terminal cost.

\subsection{Procedure}

Before the study, participants received information about the purpose of the study and data collection. They provided informed consent, acknowledging that their participation was voluntary and that they would not receive descriptions of the robot algorithms until the end of the study.
They were then brought into the motion capture area and instructed to wear a helmet to be tracked by the mocap system. They were instructed to hold the object with both hands, keeping it parallel to the ground, and to collaborate with the robot in whichever way felt natural to them, including moving around the robot. To get familiarized with the interaction and the task, participants completed three practice rounds in which the robot was teleoperated before the main experiment.

For the main portion of the experiment, participants experienced each condition (a different algorithm embodied on the robot) three consecutive times. After each condition, participants were informed of the number of times they succeeded at the task with the robot and directed to respond to a questionnaire about their latest experience.
To account for ordering effects, we randomized items in both questionnaires every time they were presented.
Additionally, we performed counterbalancing of the conditions by randomly assigning participants to one of six possible orderings of the three conditions.

\begin{table}
\centering
\caption{Summary of objective metrics. Success rate is calculated out of 72 total trials for each algorithm. Mean and standard deviation of Time and Acceleration consider only successful trials.\label{table:h1}}
\resizebox{\linewidth}{!}{
\begin{tabular}{llll}
     \toprule
     Metric                              & IC-MPC        & Vanilla-MPC  & VRNN \\
     \midrule
     Success rate (\%) $\uparrow$        & \textbf{98.6} & 88.9         & 51.4 \\
     Completion time (s) $\downarrow$    & 28.54 (1.26)  & 28.86 (3.68) & \textbf{24.98 (1.67)}  \\
     Acceleration ($m/s^2$) $\downarrow$ & 0.67 (0.15)   & 0.69 (0.17)  & 0.71 (0.14) \\
     \bottomrule
\end{tabular}
}
\end{table}

\subsection{Analysis}

We found that objective and subjective metrics did not uniformly pass the Shapiro-Wilk test of normality. Thus, for consistency, we use the non-parametric Friedman test to detect effects of the robot algorithms on dependent variables and the non-parametric paired Wilcoxon signed-rank test with Holm-Bonferroni corrections~\citep{holm1979simple} for post-hoc pairwise comparison tests. Effect sizes are reported using Kendall's coefficient of concordance (denoted as $W_k$ to disambiguate it from the test statistic of the Wilcoxon signed-rank test) and Cohen's $d$.

\textbf{H1}.
We report a summary of objective metrics for each algorithm in~\tabref{table:h1}.
IC-MPC exhibited substantially higher \emph{success rate} compared to both Vanilla-MPC and VRNN.
We found a significant effect of the robot algorithm on \emph{completion time} ($\chi^2(2)=20.61, p < 0.001, W=0.45$). Post-hoc tests indicated that, of successful trials, IC-MPC completed the task \emph{slower} compared to VRNN ($W=3.0, p<0.001, d=2.4$).
We found no significant difference in \emph{completion time} between IC-MPC and Vanilla-MPC ($W=92.0, p=0.17, d=-0.11$). Finally, of successful trials, we found no significant effect of the robot algorithm on \emph{acceleration} ($\chi^2(2)= 3.73, p=0.15, W=0.08$).
Thus, we find partial support for \textbf{H1}.

\begin{table}
\centering
\caption{Mean and standard deviation of subjective metrics.\newline$^{*}p<.05$, $^{**}p<.01$, $^{***}p<.001$\label{table:h2-pairwise}}
\resizebox{\linewidth}{!}{
\begin{tabular}{llll}
     \toprule
     Metric                   & IC-MPC               & Vanilla-MPC        & VRNN                \\
     \midrule
     Warmth~\citep{carpinella2017robotic}  $\uparrow$      & 3.44 (1.89)          & 3.12 (1.98)        & 2.99 (1.82)         \\
     Competence~\citep{carpinella2017robotic} $\uparrow$   & \textbf{6.06 (1.86)} & 5.15 (1.82)$^{*}$  & 4.02 (1.88)$^{***}$ \\
     Discomfort~\citep{carpinella2017robotic} $\downarrow$ & \textbf{2.15 (1.29)} & 2.86 (1.84)$^{*}$  & 3.22 (1.49)$^{***}$ \\
     \midrule
     Fluency~\citep{hoffman2019evaluating} $\uparrow$      & \textbf{5.73 (1.02)} & 4.64 (1.41)$^{**}$ & 3.67 (1.49)$^{***}$ \\
     \bottomrule
\end{tabular}
}
\end{table}

\textbf{H2}. We report subjective measures in~\tabref{table:h2-pairwise} and related test statistics in Table~\ref{table:h2-friedman}. Prior to testing, we evaluated Cronbach's alpha within each RoSAS subscale (\emph{warmth} $\alpha=0.90$, \emph{competence} $\alpha=0.92$, \emph{discomfort} $\alpha=0.85$) and within the \emph{Fluency} scale ($\alpha=0.95$) and found that all subscales had high internal consistency.
Individual users' responses to items within each subscale were subsequently averaged for analysis.

\begin{table}
\centering
\caption{Friedman test results on subjective measures.\label{table:h2-friedman}}
\resizebox{.7\linewidth}{!}{%
\begin{tabular}{llll}
     \toprule
     Metric         & $\chi^2(2)$ & $p$       & $W_k$ \\
     \midrule
     Warmth~\citep{carpinella2017robotic}       & 3.85        & 0.146     & 0.08   \\
     Competence~\citep{carpinella2017robotic}   & 18.86       & $< 0.001$ & 0.39   \\
     Discomfort~\citep{carpinella2017robotic}   & 15.46       & $< 0.001$ & 0.32   \\ \midrule
     Fluency~\citep{hoffman2019evaluating}      & 28.15       & $< 0.001$ & 0.59   \\
     \bottomrule
\end{tabular}
}
\end{table}

We found a significant effect of the robot algorithm on users' perception of \emph{competence}, \emph{discomfort}, and \emph{fluency}.
Post-hoc tests found that IC-MPC was judged by users as: significantly more \emph{competent} compared to Vanilla-MPC ($W=55, p=0.021, d=0.49$) and VRNN ($W=26, p<0.001, d=1.09$); significantly less \emph{discomforting} compared to Vanilla-MPC ($W=30, p=0.018, d=-0.44$) and VRNN ($W=13.5, p<0.001, d=-0.76$); a significantly more \emph{fluent} collaborator compared to Vanilla-MPC ($W=28, p=0.002, d=0.89$) and VRNN ($W=11, p<0.001, d=1.61$).
No statistically significant effect was found on users' perception of \emph{warmth}.
Thus, we find partial support for \textbf{H2}.

\begin{figure*}
  \includegraphics[width=\textwidth]{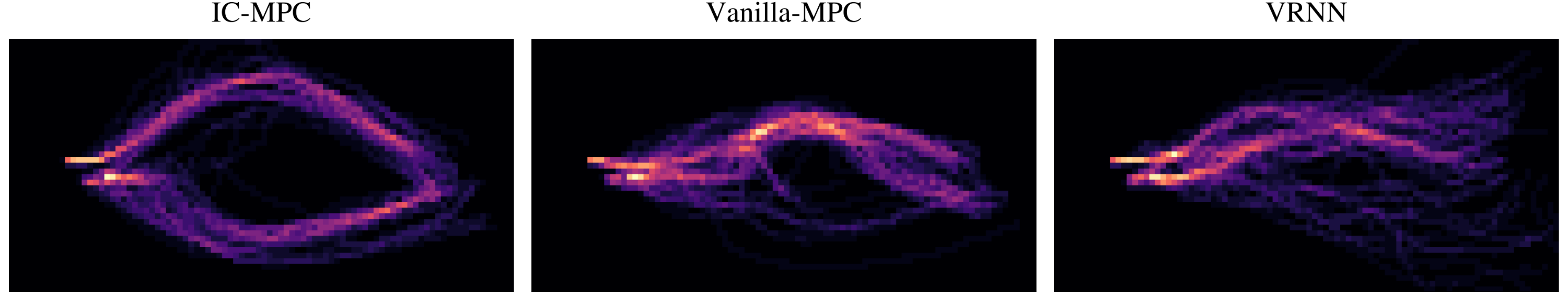}
  \caption{Spatial distribution of object trajectories within the workspace during the user study, including failure cases, itemized per algorithm. IC-MPC exhibits an almost uniform split between right and left, whereas baselines show mixed performance, including undesirable zig-zagging effects, an artifact of increased uncertainty over the unfolding traversal strategy.
  }
  \label{fig:qualitative-results}
\end{figure*}

\subsection{Discussion}

\begin{figure}
    \centering
    \includegraphics[width=\columnwidth]{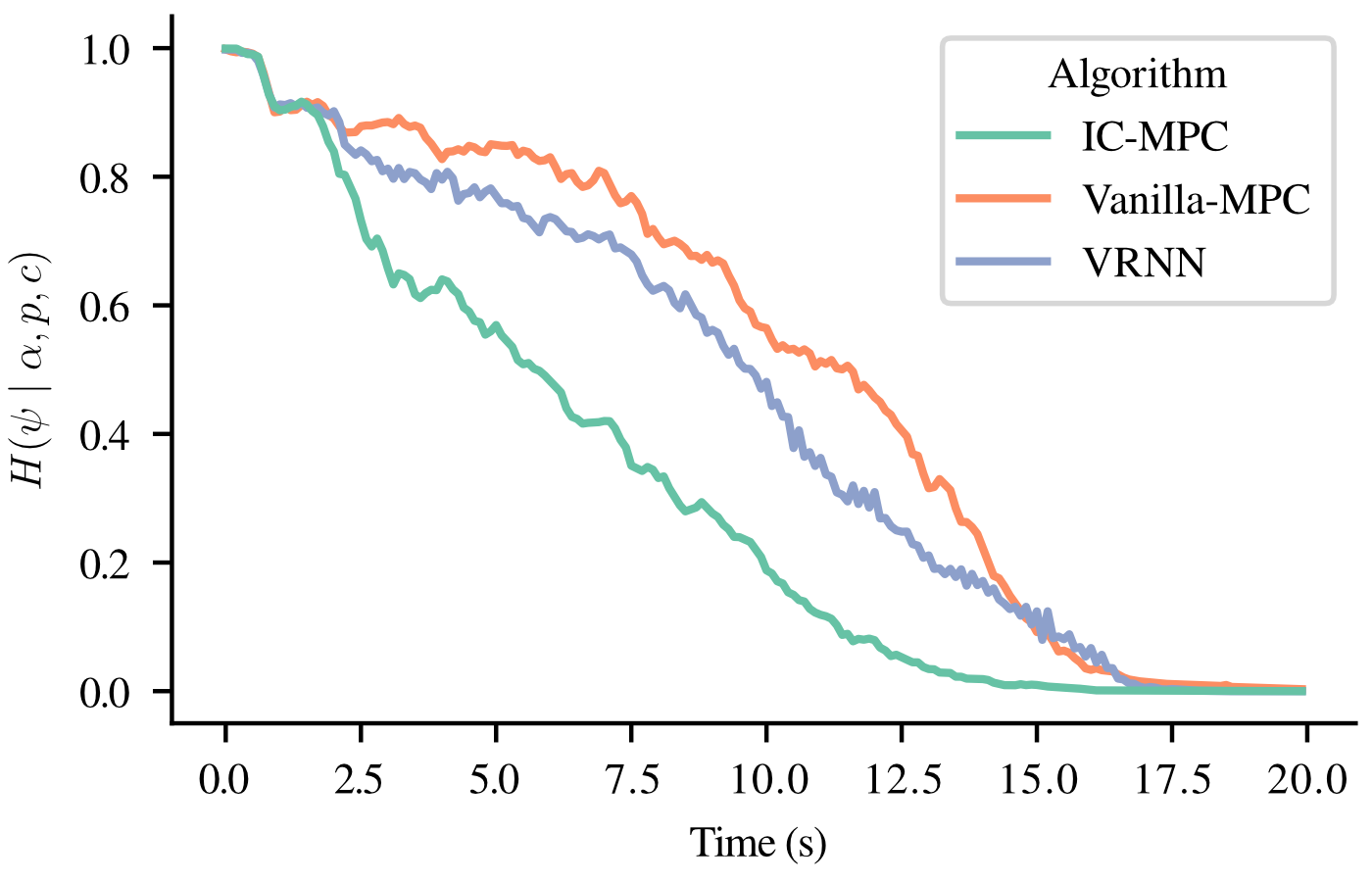}
    \caption{Entropy over the workspace traversal strategy as a proxy for strategy uncertainty, averaged across all trials for each algorithm. By directly minimizing the entropy, IC-MPC accelerates consensus on a traversal strategy. This reduces undesirable zig-zagging artifacts, present in the execution of baselines (see~\figref{fig:qualitative-results}).}
    \label{fig:post-hoc-uncertainty}
\end{figure}

Our hypotheses were motivated by a supposition that an absence of communication would create situations in which the two agents attempt to follow contradictory strategies, resulting in longer task duration, sudden movements to avoid collisions, or task failure altogether.
We believed these situations would have a negative impact on users' opinion of the robot as a teammate.
To better understand our findings, we examine motion capture data collected during the study.

The spatial distribution of object trajectories (\figref{fig:qualitative-results}) reveals that teams took \emph{wider} paths around the obstacle when the robot was running IC-MPC compared to Vanilla-MPC and VRNN.
This difference in behavior may explain our findings for \emph{success rate} and \emph{completion time}.
Without implicit communication mechanisms to resolve ambiguity or achieve consensus on traversal strategy, both baselines would often drive straight towards the goal and attempt to pass the obstacle from directions opposite the user, leading to collisions.
By acting \emph{earlier} to take \emph{wider} paths compared to Vanilla-MPC and VRNN, IC-MPC reduced uncertainty about the joint strategy faster than baselines (\figref{fig:post-hoc-uncertainty}), thereby reducing the chance of similar collisions.
As wider paths are longer than more direct paths, teams took more time on average to complete the task when the robot was running IC-MPC compared to VRNN.
Similarity between average \emph{completion time} of IC-MPC and Vanilla-MPC is coincidental: Vanilla-MPC had a tendency to slow to a near stop as it approached close to the obstacle.

We speculated that users would make sudden movements to avoid collisions caused by disagreement in joint strategy, particularly when collaborating with baselines that lack a communicative mechanism. However, in this study, we found no significant difference in the average \emph{acceleration} of the human across robot algorithms. We made two observations that could serve as possible explanations. First, the maximum speed of the Stretch RE2 (0.3 m/s) is well below users' average walking speed, which caused them to consistently alternate between taking a step and pausing to wait for the robot to cover more distance, regardless of the algorithm. Second, the slow speed of the robot gave users a relatively long time to change course to avoid a collision, if they decided to do so. Thus, \emph{sudden} movements were uncommon.

Users noticed qualitative differences among algorithms. In open-ended responses, they described VRNN as ``unpredictable'' and ``indecisive''.
One user described Vanilla-MPC as ``a bad teammate that only does what they think is right''.
Two users who interacted with IC-MPC after one or both baselines whose comments were comparative in nature wrote that IC-MPC ``felt more natural'' and that ``the collaboration on the task was a lot more seamless in this series of attempts''.

In contrast to \emph{competence}, \emph{discomfort}, and \emph{fluency}, we did not find any significant effect of robot algorithm on users' perception of \emph{warmth}.
This is not surprising, as we did not design IC-MPC or the robot to be anthropomorphic or socially expressive.
At the start of each study session, we intentionally provided the vague instruction to ``collaborate in whichever way feels natural''.
However, we received informal feedback from several participants that aspects of the interaction, including communication with the robot, \emph{did not feel natural} or \emph{intuitive}.
Participants expressed confusion about \emph{how} they could communicate with the robot and whether the robot was acknowledging, understanding, or ignoring what they were trying to communicate.
Designing the robot to be expressive may facilitate interactions that are perceived as more natural.

\section{Limitations}\label{sec:limitations}

While our decision-making framework is agnostic to the types of actions that agents are executing, in this work we considered a quasistatic setting integrating only velocities that the agents transmit to the transported object. Future work will integrate actions stemming from additional modalities, such as force, language, body posture, eye gaze, and gestures. Additionally, while our framework is not prescriptive on the space of joint strategies $\Psi$, this work emphasizes the topological relationships between the human-robot team and the obstacles around it. Future work will integrate additional strategy attributes, such as leadership~\citep{mortl2012role}, and \emph{timing} of critical maneuvers. Our control implementation was based on a flexible and practical MPC framework, but alternative approaches could be explored, such as POMDPs~\citep{nikolaidis2017human}. Our controller demonstrated practical performance assuming a simple model of human motion prediction (constant velocity); future work will integrate more fine-grained prediction models~\citep{salzmann2023hst,yasar2024posetron}. The relatively low max base speed (0.3 m/s) and payload (2 kg) of our robot might have influenced the types of interactions with users. We aspire to deploy our framework on a robot with higher payload and maximum speed to expand to tasks involving heavier objects and allow for movement across all 6 object DoFs. We also plan on expanding to more complex environments with more obstacles and dynamic agents.

\balance
\bibliographystyle{abbrvnat}
\bibliography{main.bib}

\end{document}